\documentclass[11pt,a4paper]{article}
\usepackage[hyperref]{emnlp2020}
\usepackage{times}
\usepackage{latexsym}
\usepackage{amsfonts}
\usepackage{amstext}
\usepackage{amsmath}
\usepackage{algorithm}
\usepackage{array}
\usepackage{multirow}
\usepackage{algpseudocode}
\usepackage{amsmath}
\usepackage{graphics}
\usepackage{epsfig}
\usepackage{subfigure}
\usepackage{CJKutf8}

\usepackage{microtype}

\aclfinalcopy % Uncomment this line for the final submission
\def\aclpaperid{1738} %  Enter the acl Paper ID here

\title{Coarse-to-Fine Pre-training for Named Entity Recognition}

\author{Mengge Xue$^{1,2}$ \ Bowen Yu$^{1,2}$ \ Zhenyu Zhang$^{1,2}$ \ Tingwen Liu$^{1,2}$\thanks{\hspace{0.15cm}Corresponding   Author} \ Yue Zhang$^{3}$ \and Bin Wang$^{4}$  \\
     $^1$Institute of Information Engineering, Chinese Academy of Sciences. Beijing, China \\
     $^2$School of Cyber Security, University of Chinese Academy of Sciences. Beijing, China \\
     $^3$School of Engineering, Westlake University, Hangzhou, China \\
      $^4$Xiaomi AI Lab, Xiaomi Inc., Beijing, China \\
     {\tt \{xuemengge, yubowen, zhangzhenyu1996, liutingwen\}@iie.ac.cn} \\
     {\tt yue.zhang@wias.org.cn}\qquad\quad{\tt wangbin11@xiaomi.cn}
            }

\date{}

\begin{document}
\maketitle
\begin{abstract}
More recently, Named Entity Recognition has achieved great advances aided by pre-training approaches such as BERT.
However, current pre-training techniques focus on building language modeling objectives to learn a general representation, ignoring the named entity-related knowledge. To this end, we propose a NER-specific pre-training framework to inject coarse-to-fine automatically mined entity knowledge into pre-trained models.
%Specifically, we first transfer the pre-trained model into a model for a general-typed NER task by training it with the anchor-text tagged data.
Specifically, we first warm-up the model via an entity span identification task by training it with Wikipedia anchors, which can be deemed as general-typed entities.
Then we leverage the gazetteer-based distant supervision strategy to train the model extract coarse-grained typed entities.
Finally, we devise a self-supervised auxiliary task to mine the fine-grained named entity knowledge via clustering.
%Empirical studies on three public NER datasets demonstrate the effectiveness of our proposed method. Additionally, we also exploit the positive impact of the finally fine-tuned model used as a pre-trained model for the supervised NER, establishing the state of the art on three benchmarks. We have released the source code at Github.
Empirical studies on three public NER datasets demonstrate that our framework achieves significant improvements against several pre-trained baselines, establishing the new state-of-the-art performance on three benchmarks.
Besides, we show that our framework gains promising results without using human-labeled training data, demonstrating its effectiveness in label-few and low-resource scenarios.\footnote{The source code can be obtained from https://github.com/strawberryx/CoFEE}
% We will release the source code at Github. of this paper
\end{abstract}

\section{Introduction}\label{section:introduction}
Named Entity Recognition (NER) is the task of discovering information entities and identifying their corresponding categories, such as mentions of people, organizations, locations, temporal and numeric expressions~\cite{freitag2004trained}.
It is an essential component in many applications including machine translation~\citep{babych:2003EAMT}, relation extraction~\citep{Yu:2019IJCAI}, entity linking~\citep{Xue:2019IJCAI}, and so on.

Recently, NER has seen remarkable advances with the help of pre-trained representation models, such as BERT~\cite{devlin:2019naacl} and XLNet~\cite{yang2019xlnet}.
Providing contextual representation, these pre-trained models could be easily applied to NER applications as an encoder by just fine-tuning it.
Despite refreshing the state-of-the-art performance of NER, the current pre-training techniques are not directly optimized for NER.
Typically, these models build unsupervised training objectives to capture dependency between words and learn a general language representation~\cite{tian2020skep}, while rarely considering incorporating named entity information which can provide rich knowledge for NER.
Due to little knowledge connection between NER and general language modeling, how to adapt public pre-trained models to be NER-specific remains an open problem.

To this end, injecting named entity knowledge during pre-training is a  possible solution.
However, this process of knowledge acquisition may be inefficient and expensive.
In fact, there are extensive weakly labeled annotations that naturally exist on the web yet to be explored for NER model pre-training, which are relatively easier to obtain compared with labeled data~\cite{cao:2019EMNLP}.
One can collect them from online resources, such as the Wikipedia anchors and gazetteers (named entity dictionaries).
Although automatically derived corpora usually contain massive noisy data, it still contains some extend the valuable semantic information required for NER~\cite{peng:2019acl}.
%Usually, if a human being desires to recognize a kind of named entity, first that one should know what an entity is, and then discriminate entity type.

In this paper, we propose a \emph{Coarse-to-Fine Entity knowledge Enhanced} (CoFEE) pre-training framework for NER task, aiming to gather and utilize knowledge related to named entities.
In particular, we first extract anchors from Wikipedia and use them as training corpora for entity span identification.
While anchors have no entity type information, the model could get general-typed entity knowledge from them and learn to distinguish entity words and non-entity words.
In the second phase, we use gazetteers and anchors to generate weakly labeled data for specific entity types and use it to train the model for extracting entities with coarse-grained type.
%Since the gazetteer may only cover part of entities, it is obvious that the data labeled by gazetteer contains many missing labels. With this consideration, an iterative training strategy is applied to progressively reduce the effect of false-negative labels.
%Considering that the gazetteer may only cover a part of entities, an iterative self-picking strategy is applied to progressively reduce the effect of false-negative labels.
Furthermore, another observation is that entities with the same coarse-grained type may belong to different fine-grained types.
%Although we do not know the ground truth fine-grained labels,
According to the cluster hypothesis~\cite{chapelle2009semi}, the features of entities with the same latent fine-grained label will cluster together in the semantic space.
Intuitively, mining these latent cluster structures provides auxiliary information about the coarse-grained entity type, which could be beneficial to improve the NER performance.
Based on such motivation, we finally devise a self-supervised method to exploit fine-grained type knowledge and tap the potential of weakly labeled data,  which effectively train the NER model with clustering- generated pseudo labels. 
%and prevent the model from drifting due to label noise with a variance-weighted cross-entropy loss. 
% the NER model is trained with the clustering-generated pseudo labels, which prevents the model from drifting due to label noise.
 % ÔÚ±ê×¢Êý¾Ý¼¯ÉÏxxxx
 % ÎÞÐè±ê×¢Êý¾ÝÔõÃ´Ñù
 % Ò×ÓÃÐÔ

 We conduct experiments on three realistic NER benchmarks in this paper.
 Experimental results show that the proposed CoFEE pre-training framework significantly outperforms other competitive baselines, often by large margins.
We also demonstrate that CoFEE pre-training can work well in more challenging, label-free and low-resource scenarios. 
Further ablation studies show the impact of each pre-training task in achieving these strong performance.
To the best of our knowledge, this is the first work that has tackled NER-specific representation during pre-training.

\begin{figure*}[t]
\centering
\includegraphics[width=0.98\textwidth]{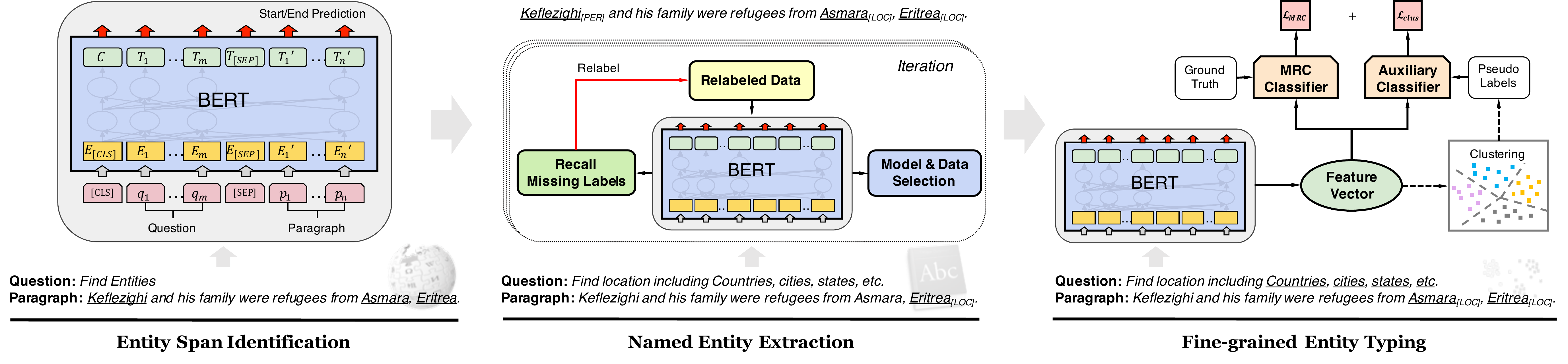}
\caption{
\label{Ourmodel_Framework_Fig}
The overall architecture of CoFEE.
In Fine-grained Entity Typing, the {\it solid line} represents the training phase and the {\it dotted line} represents the clustering phase. These two stages are iteratively done until the network converges.
}
\end{figure*}
\section{Related Work}
\paragraph{Entity Knowledge for NER.}
Recently, neural networks have been used for NER and achieved great success~\citep{Collobert:2011Natural,dos:NEW2015,Huang:2015arXiv,Ma:2016ACL}.
Specifically, various types of entity knowledge, including lexical words, gazetteers and anchors in Wikipedia have been proved to be useful for a wide range of sentiment analysis tasks.

%For supervised NER task, some researchers utilize lattice structure to incorporate the lexical information into character-based NER and avoid the segmentation error propagation of word~\cite{zhang:ACL2018}, they either design a model to be compatible with lattice input such as lattice LSTM, LR-CNN~\cite{Gui:2019IJCAI}, PLTE~\cite{Xue:2019arXiv},
%or convert lattice into graph and design graph neural networks to encode it~\citep{gui:2019emnlp,sui:2019emnlp}.
%Additionally, gazetteers have long been regarded as a useful knowledge for NER, previous methods commonly incorporated gazetteers by either using them to as handcraft features~\citep{alan:2011ACL,dominic:2018ACL} or as the auxiliary structural information for character-based NER~\citep{ding:2019ACL,liu:2019ACL}.
For supervised NER task, some researchers utilize lattice structure to incorporate the lexical information into character-based NER and avoid the segmentation error propagation of word~\cite{zhang:ACL2018,Gui:2019IJCAI,Xue:2019arXiv,gui:2019emnlp,sui:2019emnlp}.
Additionally, gazetteers have long been regarded as a piece useful knowledge for NER, previous methods commonly incorporated gazetteers by either using them as handcraft features~\citep{alan:2011ACL,dominic:2018ACL} or auxiliary structural information~\citep{ding:2019ACL,liu:2019ACL}.

For weakly supervised NER, a typical line of methods centres around transfer learning to extract source knowledge for target, such as cross-domain~\citep{yang:2017ICLR,lin:2018EMNLP,jia:2019acl} or cross-lingual~\citep{ni:2017acl,xie:2018emnlp,zhou:2019ACL}.
There are also a lot of weak labels lying on the web or gazetteers, which have not been explored.
Consequently, a number of works focus on distantly supervised methods, using anchors or gazetteers to generate data by
distant supervision~\citep{Liu:2015Information,yang:2018COLING,cao:2019EMNLP,peng:2019acl}.

\paragraph{Task Specific Pre-training.}
Unsupervised language model pre-training and task-specific fine-tuning achieve SOTA results on many NLP tasks, including NER~\citep{peters:2018naacl,devlin:2019naacl,li:2020ACL}.
Recently, with the help of automatically minded knowledge lying in the web, researchers devoted them to the pre-training models for specific tasks, including word sense disambiguation~\citep{huang:2019emnlp}, word-in-context tasks~\citep{levine:2020acl}, entity-linking and relation classification~\citep{zhang:2019ACL}, sentiment classification~\citep{tian2020skep}.

\section{Background}
In this section, we give a brief introduction to MRC-NER~\citep{li:2020ACL}, which achieves satisfying performance in NER and thus is chosen as the foundation of our work.
%Here we introduce some notations to facilitate subsequent descriptions.
Given an input paragraph $X=\{x_1, x_2, \cdots, x_n\}$ where $x_i$ denotes the i-th character, NER aims at discovering each entity $x_{start,end}$ in $X$ and identify its corresponding type $y \in Y$, where $Y$ is the set of predefined tags(e.g., PER, LOC).
$x_{start,end}=\{x_{start},x_{start+1},\cdots, x_{end-1}, x_{end}\}$ is a substring of $X$ satisfying \emph{start} $\leq$ \emph{end}.
Specifically, MRC-NER formulates NER as a machine reading comprehension (MRC) problem.
Each entity type $y$ is characterized by a natural language query $Q_y=\{q^y_1, q^y_2, ..., q^y_m\}$, and entities are extracted by answering these queries given the contexts.
For example, the task of assigning the PER label to ``[Washington] was born into slavery on the farm'' is formalized as answering the question ``Find person including fictional''.
This strategy naturally introduces the natural language query which encodes significant prior knowledge about the entity category to extract.

Formally, MRC-NER model concatenates the query $Q$ and paragraph $X$, forming string $\{[{\text{CLS}}], Q, [{\text{SEP}}], X\}$, where $[{\text{CLS}}]$ and $[{\text{SEP}}]$ are special tokens.
Then BERT~\citep{devlin:2019naacl} captures the contextual information for each token in the string via self-attention and produces the representation matrix $\textbf{H}\in \mathbb{R}^{n \times d}$ of $X$, where $d$ is the dimension of the last layer of BERT.
To extract entity spans, the representation of each word is fed to two softmax layers to predict the probability of each token being a start or end index as follows:
\begin{align}
 &P_{\text{start}}(\mathbf{y}_i^\text{s}|x_i) = softmax(\mathbf{W}_\text{s}\mathbf{h}_i+\mathbf{b}_\text{s}), \\
 &P_{\text{end}}(\mathbf{y}_i^\text{e}|\mathbf{x}_i) = softmax(\mathbf{W}_\text{e}\mathbf{h}_i+\mathbf{b}_\text{e}),
\end{align}
where $\mathbf{W}_\text{s}, \mathbf{W}_\text{e}\in \mathbb{R}^{d\times2}$ and $\mathbf{b}_\text{s}, \mathbf{b}_\text{e}\in \mathbb{R}^2$ are trainable parameters.
At training time, $S$ associated with each question $Q_y$ is paired with two label sequences $\mathbf{Y}_{\text{start}}=\{y_1^\text{s}, y_2^\text{s}, ..., y_n^\text{s}\}$ and $\mathbf{Y}_{\text{end}}=\{y_1^\text{e}, y_2^\text{e}, ..., y_n^\text{e}\}$, where $y_i^\text{s}$ ($y_i^\text{e}$) is the ground-truth label of $x_i$ being the start (end) index of a $y$-typed entity or not. The cross-entropy loss of start and end index predictions are therefore denoted as:
\begin{align}
&\mathcal{L}_{\text{start}}^{\mathcal{D}}= -\frac{1}{n}\sum_{i=1}^n y_i^\text{s} log(P_{\text{start}}(y_i^\text{s}|x_i)),\\
&\mathcal{L}_{\text{end}}^{\mathcal{D}}= -\frac{1}{n}\sum_{i=1}^n y_i^\text{e} log(P_{\text{end}}(y_i^\text{e}|x_i)),
\end{align}
where $\mathcal{D}$ denotes the training dataset.
Finally, the overall training objective to be minimized can be formulated as follows:
\begin{equation}
\mathcal{L}_{\text{MRC}}^{\mathcal{D}}=\mathcal{L}_{\text{start}}^{\mathcal{D}}+\mathcal{L}_{\text{end}}^{\mathcal{D}}.
\label{equ:mrc}
\end{equation}

\section{Methodology}

In this section, we introduce the overall framework of our coarse-to-fine pre-training. Figure~\ref{Ourmodel_Framework_Fig} gives a brief illustration, which operates in three stages as follows:
(1) Stage 1: identity entity span based on Wikipedia anchors;
(2) Stage 2: extract coarse-grained entities based on gazetteers;
(3) Stage 3: predict fine-grained entity types with a clustering-oriented self-supervised method.

\subsection{Entity Span Identification}

Pre-trained language Models such as BERT~\cite{devlin:2019naacl} and XLNet~\cite{yang2019xlnet} have been proven to capture rich language information from text.
However, as the entity information of a text is seldom explicitly studied, it is hard to expect such pre-trained general representations to capture entity-centric knowledge.
In order to better capture entity information and learn NER-specific representation, we propose the first pre-training task named Entity Span Identification (ESI).
The entity-centric knowledge is automatically mined from the large scale Wikipedia corpus.
In Wikipedia, an anchor $\langle m, e\rangle$ links a mention $m$ to an entity $e$.
Therefore, we assign an ``Entity'' tag to each anchor in the sentence and construct a General-typed weakly labeled NER dataset $\mathcal{D}_\text{g}$ without considering the entity type.
To align with MRC-NER, the question of the generated dataset is set as ``\emph{Find Entities}''.
With the general labeled data, the  MRC-NER model can be warmed-up with loss $\mathcal{L}_{\text{MRC}}^{\mathcal{D}_\text{g}}$.
By integrating the general-typed named entity knowledge into the pre-training process, the learned representation would be incorporated with the structural information of crucial importance for NER.

\begin{CJK}{UTF8}{gbsn}
\begin{table*}[h]
\centering
%\small
\begin{tabular}{|p{0.7cm}<{\centering}|p{13cm}|}
\hline
Type & Question\\
\hline
PER & 人名和虚构的人物形象 \\
\hline
ORG & 组织包括公司，政府党派，学校，政府，新闻机构 \\
\hline
LOC & 山脉，河流自然景观的地点 \\
\hline
GPE & 按照国家，城市，州县划分的地理区域 \\
\hline
HP & 找出文中的商标，包括公司，品牌 \\
\hline
HC & 找出文中的产品，包括商品，作品，食品，用品，设施，副产品，农产品，制成品，软件产品，硬件产品，资讯产品，通讯产品，通信产品，电信产品，电脑产品，手机产品，电子产品，科技产品，其他产品 \\
\hline
ORG & organization entities are limited to named corporate, governmental, or other organizational entities. \\
\hline
PRR & person entities are named persons or family. \\
\hline
LOC & location entities are the name of politically or geographically defined locations such as cities, provinces, countries, international regions, bodies of water, mountains, etc. \\
  \hline
\end{tabular}
\caption{\label{query-map}
Neural language questions for each entity type used in our model.
}
\end{table*}
\end{CJK}

\subsection{Named Entity Extraction}

% ×¢Òâµ½£¬Ô¤ÑµÁ·¹ý³ÌÖÐ½ö½öÊ¹ÓÃÁËgeneralµÄÊý¾Ý£¬²»°üÀ¨Ä¿±êÀàÐÍ
After the ESI pre-training, the model has learned to distinguish entity words and non-entity words.
Then we step into the second phase (i.e., NEE) in which the model is trained to extract typed entities with gazetteer-labeled data.
To alleviate human effort, gazetteer-based distant supervision has been applied to automatically generate labeled data and has gained successes in NER~\citep{yang:2018COLING,peng:2019acl}.
A standard strategy is to scan through the anchor text in $\mathcal{D}_\text{g}$ using the gazetteer of a given entity type $y$ and treat anchors matched with entries of the given gazetteer as the entities with type $y$.
In this way, we can obtain a specific-typed NER dataset $\mathcal{D}_\text{s}$, which is then exploited to train the MRC-NER model by optimizing $\mathcal{L}_{\text{MRC}}^{\mathcal{D}_\text{s}}$.
Besides, in order to meet the paradigm of MRC-NER, we also generate a natural language query for each entity type.
This procedure is critical since queries encode prior knowledge about labels.
%, which has a significant influence on the distantly supervised task.
Inspired by ~\citep{li:2020ACL}, we take annotation guideline notes as references to construct queries and illustrate all of the queries used in our model in Table~\ref{query-map}.
They are theoretical description of the tag categories, thus having the ability to make the model incorporate the information within the label categories unambiguously and completely. 
\iffalse
\begin{table}[t]
\centering
\begin{tabular}{|l|p{4.5cm}|}
\hline
Entity Type & Neural Language Query\\
\hline
\hline
Person & Find person including fictional. \\
\hline
Organization & Find organizations including Companies, agencies, institutions, etc. \\
\hline
Location & Find location including Countries, cities, states, mountain ranges, bodies of water, etc. \\
\hline
\end{tabular}
\caption{\label{query-map}
Examples of the queriess.
}
\end{table}
\fi

However, as most existing gazetteers only cover part of entities, the automatically derived dataset usually contains massive noisy data including missing labels, incorrect boundaries and types.
To address this issue, we propose an iterative self-picking strategy.
At the beginning (iteration 0), the model starts with training from the original noisy label set.
At the end of each iteration, the model determines the next label set by making predictions on $\mathcal{D}_\text{s}$.
Concretely, a new entity will be extracted with type $y$ if the probabilities of its start and end indices being predicted as $y$ are both greater than a picking threshold $\delta$.
In the next iteration, we use the new derived dataset as input for the model training.
Considering that we aim to recall the missing labels, we set $\delta<0.5$. 
%Obviously, if we pick the missing labels with the same $\delta$ at each iteration, excessive noise will be introduced due to its small value, such we set $\delta$ be a linear function of the number of iterations, formulated as $\delta_i=\delta_{i-1}+0.1$, where $i$ indicates the $i$-th iteration.
The model is trained until we find the best model w.r.t. the performance on the validation set.
And the final derived dataset is denoted as $\mathcal{D}_\text{s}^{\text{best}}$.

\subsection{Fine-grained Entity Typing}

NEE pre-training focuses on teaching the model named entity knowledge about coarse-grained entity types.
However, one coarse-grained entity type may be composed of a set of fine-grained entity types.
For example, the coarse-grained type \emph{Location} includes City, Country, Bodies of water, etc.
These fine-grained types can provide auxiliary information to help us understand the meaning of \emph{Location}.
%Self-labeling has efficiency in handling distantly supervised NER task, but usually brings the challenge of fitting the pseudo data distribution generated by itself.
%To effectively consume such noisy data and further improve performance, we propose to regularize our model with xxx.
%Our motivation stems from the following observation:
%after the iterative relabeling process in the previous stage, $\mathcal{D}_\text{s}^{\text{best}}$ typically includes the entities belonging to the fine-grained types described in the queries.
With this in mind, it is intuitive to group the extracted entities with a cluster miner, and use the subsequent cluster assignments as pseudo labels to mine the fine-grained NER knowledge.
%Formally, given the entity set $S_{c}=\{s_1^{\text{c}}, s^{\text{c}}_2, ..., s^{\text{c}}_m\}$ derived from the previous stage and its representation matrix $\boldsymbol{\rm{H}}_c\in \mathbb{R}^{m\times d}$, where $s^{\text{c}}_i=s_{start_i, end_i}$ and $\boldsymbol{\rm{h}}_i^{\text{c}}=avgpool([\boldsymbol{\rm{h}}_{start_i}, \boldsymbol{\rm{h}}_{start_i+1}, ..., \boldsymbol{\rm{h}}_{end_i}])$, a cluster miner is deployed to group them into pre-defined $N$ distinct clusters\footnote{Clustering has been widely studied and many approaches have been developed for a variety of circumstances. In our work, we focus on a standard clustering algorithm, k-Means.} $C_N=\{C_1, C_2, ..., C_N\}$.
One of the most well-studied clustering algorithms is k-Means, and the simplicity and efficiency have established it as a popular means for performing clustering across different disciplines.

Formally, in order to partition the entity set $E=\{e_1, e_2, \cdots, e_M\}$ in $\mathcal{D}_\text{s}^{\text{best}}$ into pre-defined $K$ distinct clusters $\{C_k\}_{k=1}^K$, k-Means minimizes the sum of the intra-cluster variances $\sum_{k=1}^K \mathcal{V}_k$, where $\mathcal{V}_k=\sum_{i=1}^M \delta_{ik}||\mathbf{e}_i-\mathbf{m}_k||^2$ and $\mathbf{m}_k=\sum_{i=1}^M \delta_{ik}\mathbf{e}_i/\sum_{i=1}^M \delta_{ik}$ are the variance and the center of the k-th cluster, respectively, $\mathbf{e}_i = sumpool([\boldsymbol{\rm{h}}_{start_i}, \boldsymbol{\rm{h}}_{start_i+1}, ..., \boldsymbol{\rm{h}}_{end_i}])$ denotes the representation of the $i$-th entity, and $\delta_{ik}$ is a cluster indicator variable with $\delta_{ik}=1$ if $e_i\in C_k$ and $0$ otherwise.
Clustering proceeds by alternating between assigning instances to their closest center and recomputing the centers, until a local minimum is reached.
The cluster assignments are used as pseudo labels to guide the transformation of $\mathcal{D}_\text{s}^{\text{best}}$ to a pseudo-labeled fine-grained dataset $\mathcal{D}_\text{c}=\{(e_i,y^\text{c}_i)\}$, where $y^\text{c}_{i}$ is the pseudo label of $e_i$
Then we can take the negative log-likelihood of the pseudo-labeled tags as the training objective:
\begin{align}
P_c(\mathbf{y}^\text{c}_i|e_i) &= softmax(\mathbf{W}_\text{c}\mathbf{e}_i+\mathbf{b}_\text{c})\\
\mathcal{L}_{\text{clus}}&= -\frac{1}{M}\sum_{i=1}^M y^\text{c}_{i} log(P_\text{c}(y^\text{c}_i|e_i))
\label{equ:clus}
\end{align}
where $\mathbf{W}_\text{c}\in\mathbb{R}^{K\times d}$ and $\mathbf{b}_\text{c} \in\mathbb{R}^{K}$ are trainable parameters, $P_\text{c}(y^\text{c}_i|e_i)$ denotes the probability of entity $e_i$ being predicted to the $y^\text{c}_{i}$-th cluster.
Recall that our purpose is to pre-train the NER model to discover typed entities belonging to $Y$ rather than fine-grained entities, so $\mathcal{L}_{\text{clus}}$ can be deemed as an auxiliary task to assist the model to mine the fine-grained NER knowledge and regularize the optimization of $\mathcal{L}_{\text{MRC}}^{\mathcal{D}_\text{s}}$.
So the training objective in this stage is defined as:
%00
\begin{equation}
\mathcal{L}_{\text{FET}}=\mathcal{L}_{\text{MRC}}^{\mathcal{D}_\text{s}^{\text{best}}}+\gamma\mathcal{L}_{\text{clus}}, 
\label{equ:third}
\end{equation}
where $\gamma$ is the trade-off parameter.

While optimizing with pseudo labels created by the cluster miner seems reasonable, the inevitable label noise caused by the clustering procedure is ignored.
%Such noisy pseudo labels substantially hinder the model¡¯s capability to further improve the classification performance on the target domain.
To this end, we propose a variance-weighted cross-entropy loss to alleviate the influence of noisy pseudo labels.
Obviously, the inverse of $\mathcal{V}_k$ ($\mathcal{V}^{-1}_k$) represents the intra-cluster compactness of the $k$-th cluster.
% and the the confidence of the $k$-th pseudo label.
If the features of instances in the $k$-th cluster are close together, $\mathcal{V}_k^{-1}$ will be large, meantime the confidence of assigning pseudo label $y_i$ to these instances should also be high and vice versa.
Thus we re-formulate Equation~\ref{equ:clus} as:
\begin{align}
\mathcal{L}_{\text{clus}} &= -\frac{1}{M}\sum_{i=1}^M \alpha_{y^\text{c}_{i}} y^\text{c}_{i} log(P_\text{c}(y^\text{c}_i|e_i)),\\
\alpha_{y^\text{c}_{i}} &= \frac{exp(\mathcal{V}_{y^\text{c}_i}^{-1})}{\sum_{k=1}^{K}exp(\mathcal{V}_n^{-1})}.
\end{align}
Finally, we iterate the above clustering-optimizing process by putting back the model to output new representations, generate new pseudo labels $\mathcal{D}_\text{c}$ and start the next iteration.

\begin{algorithm}[t]
\small
\caption{Coarse-to-fine Pre-training }
\label{algo:trainStrategy}
\begin{algorithmic}[1]
\Require Wikipedia corpus; 
\Require Specific typed gazetteers;
\Require Specific typed validation data $\mathcal{D}_\text{s}^{val}$;
\Require Initialize Model Parameters $\theta$ with BERT.
%Specific typed validation data ;
%General typed data $\mathcal{D}_\text{g}=\{(S,\tilde{Y}^g_{start},\tilde{Y}^g_{end})\}$;
%Specific typed data $\mathcal{D}_\text{s}=\{(S,\tilde{Y}^s_{start},\tilde{Y}^s_{end})\}$;
%Cluster data $\mathcal{D}_\text{c}=\{(S_c,\tilde{Y}_c)\}$;
%$best\_acc=0$;
%\Ensure
%The best model $\theta_{best}$;
%\State \textbf{Initialize:} Model Parameters $\theta \leftarrow$ BERT parameters;

\State Construct $\mathcal{D}_\text{g}$ based on Wikipedia anchors
\For{$epoch \leftarrow 1$ to $e_1$} \algorithmiccomment{Stage 1.}
  \State Update $\theta$ w.r.t. $\mathcal{L}_{\text{MRC}}^{\mathcal{D}_\text{g}}$
%  \If{acc($\theta_g$,$\mathcal{D}_\text{g}^{val}$) $>$ acc($\theta_g$, $\mathcal{D}_\text{g}^{val}$)}
%      \State $\theta_{best}$ $\leftarrow$ $\theta_g$
%  \EndIf
\EndFor
%\State $\theta_s$ $\leftarrow$ $\theta_{best}$
\State Construct $\mathcal{D}_\text{s}$ by matching $\mathcal{D}_\text{g}$ to gazetteer.

\For{$epoch \leftarrow 1$ to $e_2$} \algorithmiccomment{Stage 2.}
  \State Update $\theta$ w.r.t. $\mathcal{L}_{\text{MRC}}^{\mathcal{D}_\text{s}}$.
  \If{$score(\theta,\mathcal{D}_\text{s}^{\text{val}})$ $>$ $best\_score$}
      \State $\theta_{\text{best}}$ $\leftarrow$ $\theta$; $\mathcal{D}_\text{s}^{\text{best}}$ $\leftarrow$ $\mathcal{D}_\text{s}$; \State $best\_score=score(\theta,\mathcal{D}_\text{s}^{\text{val}})$
  \EndIf
%  \For {$(s_i,\tilde{y}_i^s, \tilde{y}_i^e) in D_s$}
%    \State $\tilde{y}_i^s=0$; $\tilde{y}_i^e=0$
%    \If{$p_{start}(\tilde{y}_i^s|s_i)$ $>$ $\delta$ and $p_{end}(\tilde{y}_i^e|s_i)$ $>$ $\delta$}
%      \State $\tilde{y}_i^s=1$; $\tilde{y}_i^e=1$
%    \EndIf
%  \EndFor
  \State Re-label $\mathcal{D}_\text{s}$ with $\theta$.

\EndFor
\State $\theta$ $\leftarrow$ $\theta_{\text{best}}$,  $best\_score \leftarrow 0$
\State Construct $\mathcal{D}_\text{c}$ by clustering entities in $\mathcal{D}_\text{s}^{\text{best}}$.
\For{$epoch \leftarrow 1$ to $e_3$} \algorithmiccomment{Stage 3.}
  \State Update $\theta$ w.r.t. $\mathcal{L}_{\text{FET}}$
  \If{$score(\theta,\mathcal{D}_\text{s}^{val}) > best\_score$}
      \State $\theta_{\text{best}}$ $\leftarrow$ $\theta$;
      \State $best\_score=score(\theta,\mathcal{D}_\text{s}^{\text{val}})$
  \EndIf
\State Re-cluster entities in  $\mathcal{D}_\text{s}^{\text{best}}$ and construct new $\mathcal{D}_\text{c}$
\EndFor
\State \Return $\theta_{\text{best}}$
\end{algorithmic}
\end{algorithm}

\subsection{Algorithm Workflow}

In this subsection, we introduce the overall procedure of our framework. Algorithm~\ref{algo:trainStrategy} gives the scratch.
First, we construct general-typed NER data $\mathcal{D}_\text{g}$ based on Wikipedia anchors, and pre-train the model to extract general typed entities with loss $\mathcal{L}_{\text{MRC}}^{\mathcal{D}_\text{g}}$.
Then we leverage the gazetteer-based distant supervision strategy to construct a specific-typed NER dataset $\mathcal{D}_\text{s}$, and propose an iterative self-picking method to alleviate the data missing problem.
In each iteration, the model is optimized to fit the data labeled by the previous iteration.
When the performance on the validation set starts to decline, the iteration is ended and the best-performed model is passed to the third stage, where a cluster miner is deployed to group the entities extracted from the second stage into fine-grained types, and the model is trained to simultaneously distinguish fine-grained entities and extract specific-typed entities.
%The iterative re-labeling method is also utilized to gradually refine the fine-grained pseudo labels.
Also, we iteratively cluster the features from the last iteration to gradually refine the fine-grained pseudo labels for current.

\section{Experiments}
We evaluate the CoFEE framework under two settings: (i) supervised setting (ii) weakly supervised setting. In the supervised setting, the pre-trained model is fine-tuned on human-labeled datasets while in the weakly-supervised setting, the model pre-trained with CoFEE is directly applied to perform NER without fine-tuning. Next, we describe these experiments in detail.
%\footnote{Hyper-parameter settings are listed in Appendix.A.}

\subsection{Datasets}
Our experiments are conducted on three benchmarks.
(1) \textbf{Chinese Ontonotes 4.0} consists of newswire text and published by~\citeauthor{Ralph:2011}~\shortcite{Ralph:2011}.
It is annotated by four types: PER (Person), ORG (Organization), GPE (Geo-Political Entity) and LOC (Location) for Chinese named entity.
It contains 15.7k sentences for training and 4.3k for testing.
(2) \textbf{E-commerce} is a Chinese NER dataset collected from the e-commerce domain and released by~\citeauthor{ding:2019ACL}~\shortcite{ding:2019ACL}.
It is annotated by PROD (product) and BRAN (brand) types.
%, which are more challenging than traditional PER and ORG types.
The training and test datasets contain 273k and 53k lines, respectively.
(3) \textbf{Twitter} is an English NER dataset~\cite{Zhang:2018AAAI}, following \cite{peng:2019acl}, we only use textual information to perform NER and make entity detection on PER, LOC and ORG.
It contains 4,000 tweets for training and 3,257 tweets for testing.

\subsection{Pre-training Corpora}
\label{sec:pre-cor}

\paragraph{Wikipedia.}
We use 20200401 Chinese and English Wikipedia dumps\footnote{https://dumps.wikimedia.org/zhwiki/20200401/zhwiki-20200401-pages-articles.xml.bz2}\footnote{https://dumps.wikimedia.org/enwiki/20200401/enwiki-20200401-pages-articles.xml.bz2}
 for data construction, where we set the max sentence length as 250 and remove the sentences which contain three or fewer anchors.
The resulting Chinese corpora contains 1,116,514 sentences and 6,383,142 anchors (entity mentions), and the English corpora contains 3,911,059 sentences and 37,755,176 anchors.
 \paragraph{Gazetteer.}
 For Chinese PER, ORG, GPE, and LOC, we collect the gazetteers from the crowdsource dictionaries used by Chinese Input Method "Sougou"\footnote{https://pinyin.sogou.com/dict/}, which contain 2,314 person names, 2,649 organization names, 895 geopolitical entities, and 628 location names.
For Chinese PROD and BRAN, we use the gazetteers provided by \citeauthor{ding:2019ACL}\shortcite{ding:2019ACL}, which contain 628 brand names and 2,974 product names.
 For English PER, ORG and LOC, we collect the gazetteers using the method released by \citeauthor{peng:2019acl}\shortcite{peng:2019acl}, which contain 2,795 person names, 1,825 organization names and 1,408 location names.

\subsection{Baselines}
We chose two types of baselines: supervised methods and the weakly supervised methods.
We call our proposed CoFEE pre-training framework with MRC-NER backbone as CoFEE-MRC. In addition, to demonstrate the model-agnostic and generic property of CoFEE, we also implemented another competitive baseline by replacing the MRC-NER backbone with a widely used BERT model~\cite{devlin:2019naacl} without any change in the training procedure, denoted as CoFEE-BERT.
We used open-source release of https://github.com/huggingface/transformers.

\paragraph{Supervised Setting.}
We fine-tune CoFEE-MRC and CoFEE-BERT on supervised NER data and compare with the following baselines to learn how improvement can be achieved for supervised models.
\textbf{BiLSTM-CRF}~\cite{Huang:2015arXiv} is a classical neural-network-based baseline for NER, which usually achieves competitive performance in supervised NER.
\textbf{BERT-Tagger}~\citep{devlin:2019naacl} uses the outputs from the last layer of model BERT$_{base}$ as the character-level enriched contextual representations to make sequence labeling.
\textbf{MRC-NER}~\citep{li:2020ACL} formulates NER as a machine reading comprehension task and uses BERT as the basic encoder.
%\textbf{CoFEE-BERT} and \textbf{CoFEE-MRC} utilize BERT and BERT-MRC model to fine-tune the pretrained model CoFEE respectively.
%\subsection{Training Details}

\paragraph{Weakly Supervised Setting.}
We investigate the effect of CoFEE-MRC for solving the NER task without any human annotations, and compare the model to some weakly supervised NER models.
For fair comparison, we implemented baselines with the same gazetteers constructed in Section~\ref{sec:pre-cor}.
%\citet{peng:2019acl} formulates NER as a positive-unlabeled learning problem by using unlabeled training data and gazetteer.
\textbf{Gazetteer Matching} applies the constructed gazetteers to the test set directly to obtain entity mentions with exactly the same surface name.
By comparing with it, we can check the improvements of neural models over the distant supervision itself.
%\textbf{Soft} \cite{jie:naacl2019} exploits a $k$-fold cross-validation fashion to fit a reasonable label distribution for NER with incomplete annotations. We re-implemented it based on the BERT encoder.
\textbf{MRC-NER} uses the MRC-NER backbone to perform weakly supervised NER task with gazetteer labeled training data.
Furthermore, we explore the influence of our proposed pre-training tasks by removing entity span identification pre-training (\textbf{-ESI}) and fine-grained entity typing pre-training (\textbf{-FTP}) from CoFEE-MRC.

%MRC-WL with entity span identification pre-training (\textbf{+ESI}) and named entity extraction pre-training (\textbf{+NEE}).

\subsection{Hyper-parameter settings}
We use the BertAdam as our optimizer, all of the models are implemented under PyTorch using a single NVIDIA Tesla V100 GPU, we use ”bertbase-chinese” and ”bert-base-cased” as our pretrained models for Chinese and English language, the number of parameters is same to these pretrained models in addition to two binary classifier. For each training stage, we vary the learning rate from $1e-6$ to $1e-4$.
In NEE stage, we select the best trade-off $\delta$ from $0.1$ to $0.5$ with an incremental $0.1$. In FET stage, we choose the number of clusters $K$ from $\{K\!-\!2,K\!-\!1,K,K\!+\!1,K\!+\!2\}$ if we set $K$ as the categories of fine-grained entity. For all these hyper-parameters, we select the best according to the F1-score on the dev sets.

  \begin{table}[t]
 \centering
 \small
 \begin{tabular}{|p{4.35cm}p{0.45cm}p{0.45cm}p{0.6cm}|}
 \hline
  \multicolumn{4}{|c|}{\textbf{Chinese OntoNotes 4.0}}\\
  \hline
  Model                                   & P & R & F1 \\
  \hline
  BiLSTM-CRF~\citep{Huang:2015arXiv}      & 72.0             & 75.1            & 73.5          \\
  BERT~\citep{devlin:2019naacl}           & 78.01            & 80.35           & 79.16       \\
   MRC-NER~\citep{li:2020ACL}             & \textbf{82.98}	         & 81.25           & 82.11       \\
   \hline
  CoFEE-BERT                               & 80.27	         & 80.64	       & 80.46        \\
  %\multicolumn{4}{|r|}{\textbf{(+2.71)}}        \\
%  ~&~&~&\textbf{+2.71}\\
  CoFEE-MRC                                & 82.5	 & \textbf{82.78}   & \textbf{82.64}        \\
  %\multicolumn{4}{|r|}{\textbf{(+3.46)}}        \\
%  ~&~&~&\textbf{+3.46}\\
  \hline
  \hline
  \multicolumn{4}{|c|}{\textbf{E-commerce}}\\
  \hline
  Model               & P & R & F1 \\
  \hline
  BiLSTM-CRF~\citep{Huang:2015arXiv}      & 71.1             & 76.1            & 73.6          \\
  BERT~\citep{devlin:2019naacl}           & 77.06            & \textbf{80.65}           & 78.81       \\
   MRC-NER~\citep{li:2020ACL}             & 79.47	         & 78.3            & 78.88       \\
   \hline
  CoFEE-BERT                               & 79.13	         & 80.34	       & \textbf{79.73}        \\
  %\multicolumn{4}{|r|}{\textbf{(+1.87)}}        \\
%  ~&~&~&\textbf{+1.87}\\
  CoFEE-MRC                                & \textbf{80.26}	 & 78.88   & 79.56        \\
  %\multicolumn{4}{|r|}{\textbf{(+1.88)}}        \\
%  ~&~&~&\textbf{+1.88}\\
  \hline
  \hline
  \multicolumn{4}{|c|}{\textbf{Twitter}}\\
  \hline
  Model               & P & R & F1 \\
  \hline
  BiLSTM-CRF~\citep{Huang:2015arXiv}      & --               & --              & 65.32          \\
  BERT~\citep{devlin:2019naacl}           & 69.83            & 69.35           & 69.59       \\
  MRC-NER~\citep{li:2020ACL}             & 72.06	         & 70.83           & 71.44       \\
  \hline
  CoFEE-BERT                               & 75.17	         & 71.17	       & 73.11        \\
  %\multicolumn{4}{|r|}{\textbf{(+4.36)}}        \\
%  ~&~&~&\textbf{+4.36}\\
   CoFEE-MRC                                & \textbf{75.89}	 & \textbf{71.93}  & \textbf{73.86}        \\
  %\multicolumn{4}{|r|}{\textbf{(+5.76)}}        \\
%  ~&~&~&\textbf{+5.76}\\
  \hline
  \end{tabular}
 \caption{\label{supervised-Result}
Model performance (\%) for supervised NER on three benchmark datasets. Bold marks highest number among all models.
}
\end{table}

\begin{table}[t]
\renewcommand\arraystretch{1.1}
 \centering
 \small
 \begin{tabular}{|llll|}
 \hline
  \multicolumn{4}{|c|}{\textbf{Chinese OntoNotes 4.0}}\\
  \hline
  Model                  & P & R & F1 \\
  \hline
  Matching               & 28.29         & 40.95      & 33.46          \\
  %Soft\cite{jie:naacl2019}   & 38.30        & 43.48      & 40.72          \\
  MRC-NER~\citep{li:2020ACL}            & 44.85        & 33.06      & 38.06       \\
  \hline
   CoFEE-MRC                   & \textbf{48.01}        & \textbf{41.22}      & \textbf{44.36}       \\
  \;\;-FET               & 48.0        & 39.32      & 43.23       \\
   \;\;-FET-ESI                   & 48.19        & 30.64      & 40.3       \\

    \hline
  \hline
  \multicolumn{4}{|c|}{\textbf{E-commerce}}\\
  \hline
  Model                  & P & R & F1 \\
  \hline
  Matching               & 38.94        & 38.34      & 39.14          \\
  %Soft\cite{jie:naacl2019}  & 45.54        & 34.17      & 39.04          \\
  MRC-NER~\citep{li:2020ACL}            & 54.84        & 22.78      & 32.19       \\
    \hline
    CoFEE-MRC                    & 50.27        & \textbf{53.22}      & \textbf{51.7}       \\
  \;\;-FET               & 52.42         & 42.88      & 47.17       \\
  \;\;-FET-ESI                   & \textbf{55.03}        & 38.03      & 44.98       \\
    \hline
  \hline
  \multicolumn{4}{|c|}{\textbf{Twitter}}\\
  \hline
  Model                  & P & R & F1 \\
  \hline
  Matching               & 28.29        & 24.58      & 26.30          \\
  %Soft\cite{jie:naacl2019}  & 45.73           & 36.95         & 40.87          \\
  MRC-NER~\citep{li:2020ACL}            & 52.07        & 45.59      & 48.62       \\
 \hline
   CoFEE-MRC                    & \textbf{56.44}        & \textbf{52.81}      & \textbf{54.56}       \\
   \;\;-FET               & 54.92        & 51.35      & 53.07       \\
    \;\;-FET-ESI                   & 56.06        & 47.28      & 51.3       \\
  \hline
  \end{tabular}
 \caption{\label{Weakly-supervised-Result}
Model performance (\%) for weakly supervised NER on three benchmark datasets. Bold marks highest number among all models.
}
\end{table}

\begin{figure*}[htbp]
\centering
\subfigure[]{
\label{datasize_weakly}
\includegraphics[width=0.22\textwidth]{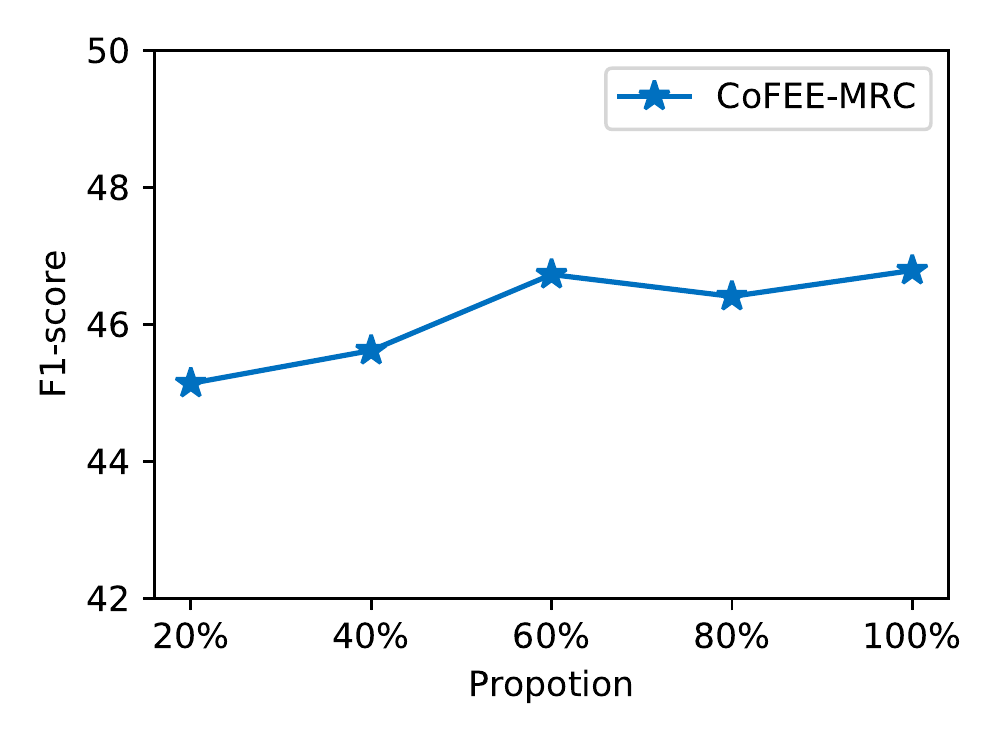}
}
\hspace{.07in}
\subfigure[]{
\label{datasize_supervised}
\includegraphics[width=0.22\textwidth]{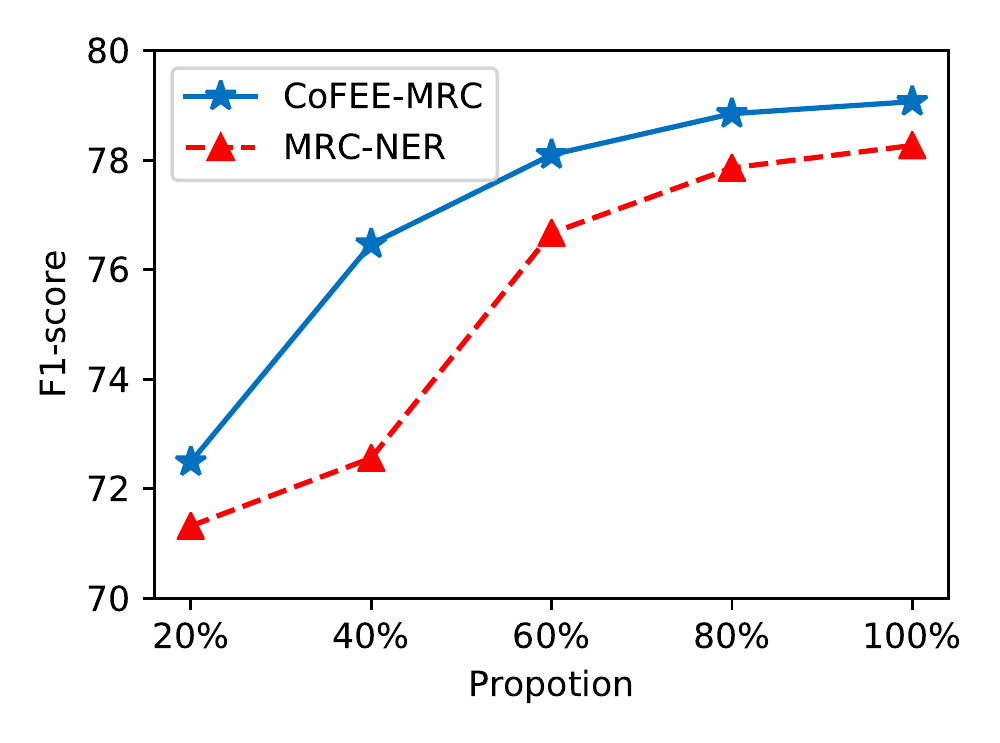}
}%
\hspace{.07in}
\subfigure[]{
\label{delta}
\includegraphics[width=0.22\textwidth]{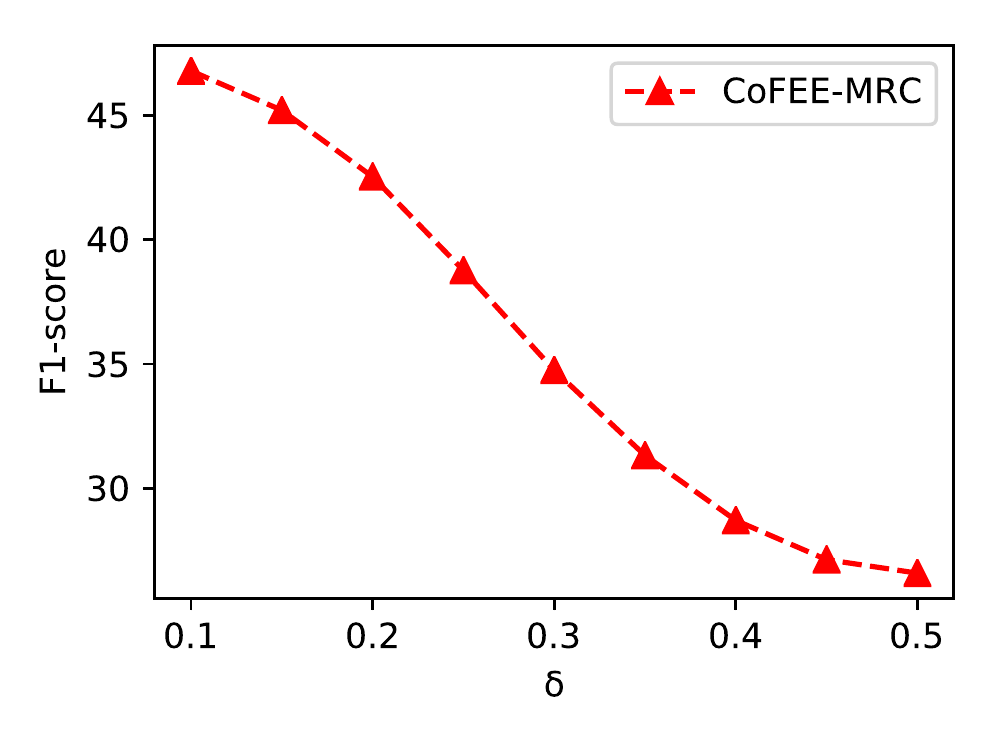}
}%
\hspace{.07in}
\subfigure[]{
\label{K}
\includegraphics[width=0.22\textwidth]{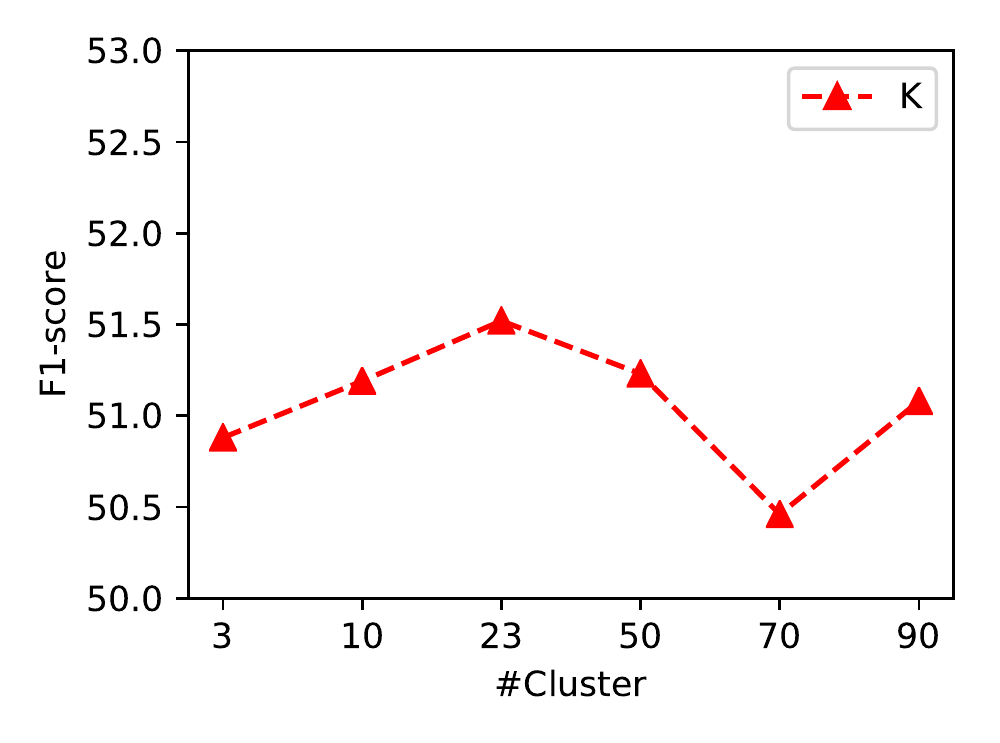}
}%
\centering
\caption{
%Impact of hyper parameters on the Weibo dev set.
(a) Impact of pre-training data size on the weakly supervised setting;
and (b) Impact of fine-tuning data size on the supervised setting;
and (c) Impact of picking rate $\delta$;
and (d) Impact of cluster size $K$.
% Experimental Results on Weibo validation set
}
\end{figure*}

\begin{figure}[htbp]
\centering
\subfigure[The number of named entities in gazetteers.]{
\label{dict_size_stastic}
\includegraphics[width=0.21\textwidth]{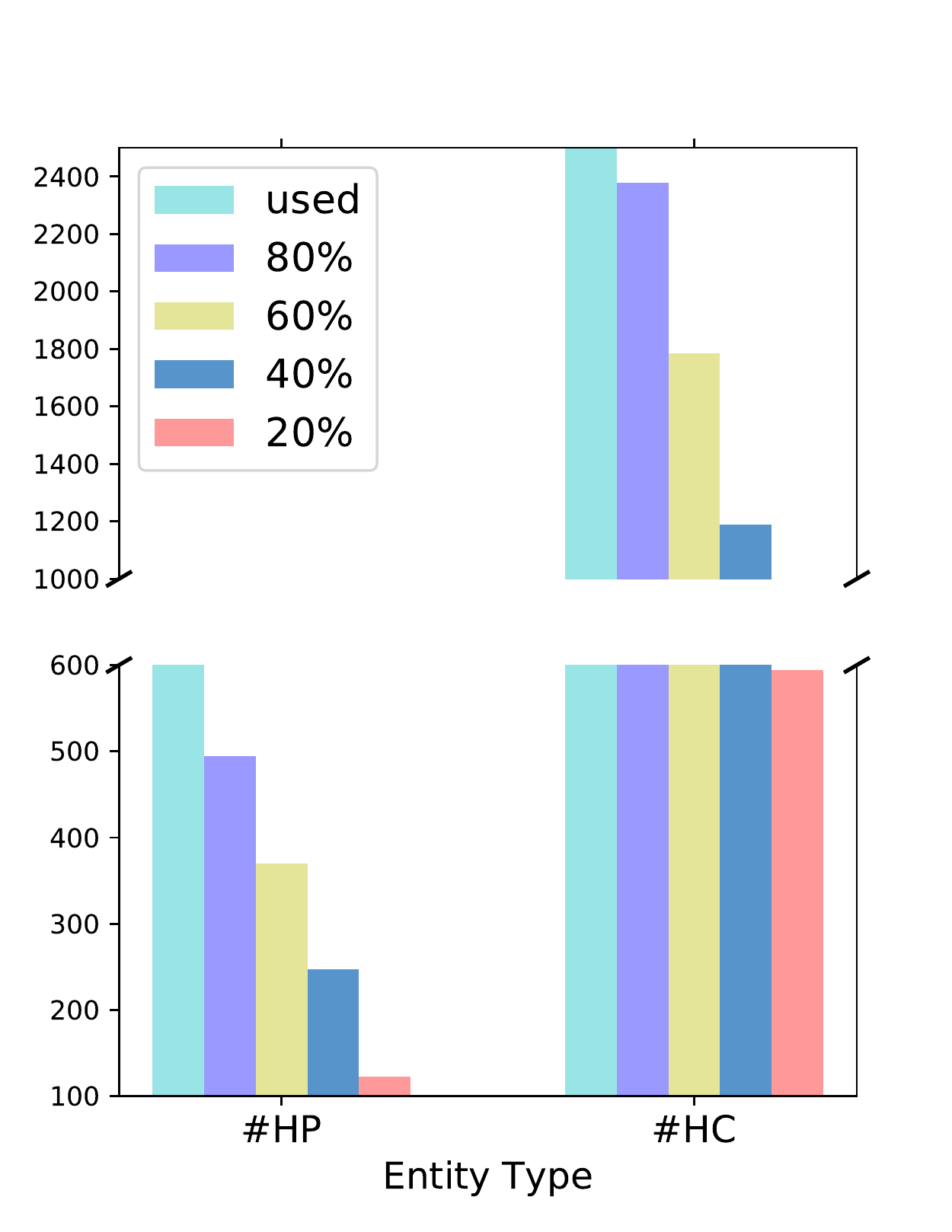}
}
\hspace{.07in}
\subfigure[The number of training sentences and the model performance.]{
\label{dict_size_performance}
\includegraphics[width=0.21\textwidth]{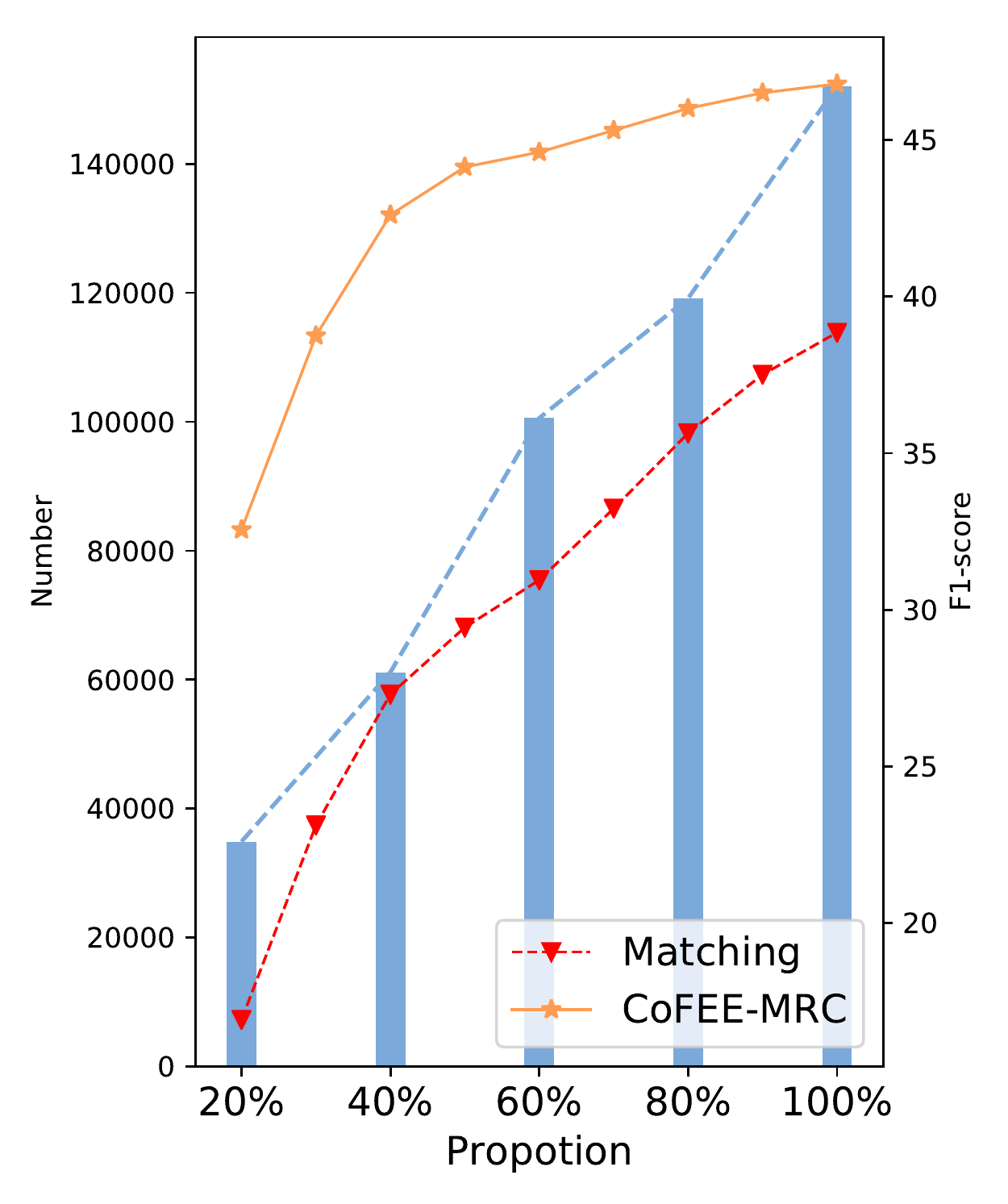}
}
\caption{\label{dict_size}
 Statistic information and model performance with gazetteers of different sizes on the Weibo dev set.
}
\end{figure}

\subsection{Evaluation}

Following the evaluation metrics in previous work ~\citep{li:2020ACL}, we apply the entity-level (exact entity match) standard micro Precision (P), Recall (R), and F1 score to evaluate the results.

\subsection{Overall Performance}
 Table~\ref{supervised-Result} contains results for models tuned on human-labeled NER data.
 We can observe that our CoFEE-MRC pre-training performs remarkably better than MRC-NER, establishing an impressive new state-of-the-art for supervised NER on OntoNotes and Twitter of $82.64\%$ and $73.86\%$, respectively.
 CoFEE-BERT also significantly improves the performance compared with BERT and achieves a new SOTA for supervised NER on E-commerce of $79.73\%$,
 which confirms the model-agnostic property of our CoFEE pre-training framework.
 Please note that the results of MRC-NER on OntoNotes have a few concerns need to be addressed. MRC-NER set the max sentence length as 77, which is far less than the true maximum length of the dataset. While in our method, we promise that the maximum length is more than 100.

Table~\ref{Weakly-supervised-Result} reports the results of our models against to baselines under the weakly supervised setting.
We can find that:
\textbf{1)} Gazetteer Matching performs quite poorly and the capability of this method is strongly influenced by the size of the gazetteers.
For OntoNotes, the coverage of the large scale gazetteer is almost $40\%$, but also its huge size causes the low precision.
For Twitter, the recall value is about $14\%$ due to its limited gazetteers.
\textbf{2)} If we directly use MRC-NER to perform weakly supervised NER task with gazetteer labeled data, the model achieves a degree of improvement but is still inaccurate due to the distantly labeled data.
\textbf{3)} CoFEE-MRC achieves the state-of-the-art F1 score on all three benchmarks, which confirms the validity of our proposed CoFEE pre-training framework.
\textbf{4)} FET pre-training task brings performance improvements,
 which verifies the effectiveness of the introducing fine-grained named entity knowledge.
\textbf{5)} ESI pre-training further improves the performance,
 which demonstrates the necessity to warm-up the pre-trained language model using general-typed named entity knowledge.

\section{Analysis}

\subsection{Impact of Data Size}\label{section_datasize}
We analyze the influence of reducing the amount of pre-training data and fine-tuning data.
%We sample sentences from the pre-training corpus and fine-tuning data, train models on the selected sentences, and evaluate them on the dev set of xxx.
The results on the dev set of E-commerce are shown in Figure~\ref{datasize_weakly} and \ref{datasize_supervised}, respectively.
%We investigate the effectiveness of different sizes of distantly supervised texts.
%We sample sentences uniformly random from the training corpus and then evaluate our model trained on the selected sentence.
%Figure~\ref{datasize_weakly} and \ref{datasize_supervised} illustrates the results of this study on distantly supervised texts and also annotated corpus, respectively.
From Figure~\ref{datasize_weakly}, we can observe that increasing the size of the pre-training data will improve the performance generally, but the improvement tends to flatten out with $60\%\sim80\%$ data.
We suppose that this is because of the number of unique patterns, the influence of the training data size has its local minimum and maximum critical point.
From Figure~\ref{datasize_supervised}, we see that knowledge enhanced pre-training is more effective for low-resource cases, where there is a larger gap in performance between our CoFEE-MRC and MRC-NER. Besides, the performance of CoFEE pre-training is more stable as data scale changes.
This further demonstrates that our CoFEE pre-training framework can significantly reduce human efforts to create NER taggers.

\subsection{Impact of Picking Rate}
We then evaluate the influence of the value and variation of our picking rate $\delta$.
From Figure \ref{delta}, we can see that setting a lower picking rate to recall more named entities can indeed improve a great performance for the model and gives the highest result with $\delta_0=0.1$.
% If we set $\delta$ as 0.4 or higher, the model achieve its best performance at 0-th iteration, that is to say, the self-picking schema with higher filter rate is useless.
%Besides, our dynamic setting which makes $\delta$ be a linear function of the iteration achieves better performance and gives the highest result with $\delta_0=0.2$.

\subsection{Impact of Gazetteer Size}

%\begin{figure}[htbp]
%\centering
%\subfigure[The number of named entities.]{
%\label{dict_size_stastic}
%\includegraphics[width=0.21\textwidth]{dict_size.pdf}
%}
%\hspace{.07in}
%\subfigure[The number of training sentences and the performance of the model.]{
%\label{dict_size_performance}
%\includegraphics[width=0.21\textwidth]{dict_size_result.pdf}
%}
%\caption{\label{dict_size}
% Statistic information and model performance using different dictionaries with different sample ratio.
%}
%\end{figure}
We further explore the change of the training data and performance when we use gazetteers of different sizes.
In particular, we used 20\%, 40\%, 60\%, 80\% and 100\% of the original gazetteers to construct pre-training corpora.
Statistical information of each resultant gazetteer is illustrated in Figure~\ref{dict_size_stastic}, and the model performance on the E-commerce dev set with these gazetteers is demonstrated in Figure~\ref{dict_size_performance}.
We can observe that increasing the size of gazetteers will generally improve the performance of our proposed CoFEE-MRC model and the performance growths in line with the performance of ``Matching'', indicating that in addition to the gazetteer size, matching degree also has a crucial influence on the model performance. 

\subsection{Impact of Cluster Size}
The proposed CoFEE framework does require a cluster size $K$ as the scope for pseudo labels.
One may wonder whether the choice of $K$ has a significant influence on the final results.
In this subsection, we vary $K$ from 4 to 90 and report the  F1 score of CoFEE-MRC on the E-commerce dev set.
As shown in Figure~\ref{K}, the best performance is obtained when $K$ is exactly set as the number of fine-grained entity types described in the queries (23), indicating that our CoFEE pre-training can leverage this information as useful prior knowledge.
Thanks to the self-supervised learning schema, when we very from 3 to 90, the model achieves stable F1 score and is not sensitive to the choice of $K$.
The results also further indicate the applicability of the proposed framework when being applied to a new kind of named entity where the number of fine-grained entity types is not available in advance. 
We can safely assign a larger value than needed and the model is still robust.

\section{Conclusion}
%In this paper, we propose Coarse-to-Fine Entity knowledge Enhanced Pre-training for named entity recognition, which consumes three kinds of entity knowledge with different granularity levels.
We investigated coarse-to-fine entity knowledge enhanced pre-training for named entity recognition, which integrates three kinds of entity knowledge with different granularity levels.
Though conceptually simple, our framework is highly effective and easy to implement.
On three popular NER benchmarks, we found consistent improvements over both state-of-the-art supervised and weakly-supervised methods.
Further analysis verifies the necessity of utilizing NER knowledge for pre-training models.

\section*{Acknowledgements}
We would like to thank the anonymous reviewers for their insightful comments and suggestions.
This research is supported by the National Key Research and Development Program of China (grant
No.2016YFB0801003) and the Strategic Priority
Research Program of Chinese Academy of Sciences (grant No.XDC02040400).

\bibliography{anthology,emnlp2020}
\bibliographystyle{acl_natbib}
\end{document}